\documentclass{article}

\usepackage{arxiv}
\usepackage{amsmath}

\usepackage[utf8]{inputenc} 
\usepackage[T1]{fontenc}    
\usepackage{hyperref}       
\usepackage{url}            
\usepackage{booktabs}       
\usepackage{amsfonts}       
\usepackage{nicefrac}       
\usepackage{microtype}      
\usepackage{cleveref}       
\usepackage{lipsum}         
\usepackage{graphicx}
\usepackage[square,sort,comma,numbers]{natbib}
\usepackage{doi}
\usepackage{textcomp}
\usepackage{booktabs}
\usepackage{subcaption}
\usepackage{xcolor}
 \usepackage{multirow}

\title{Unsupervised training of keypoint-agnostic descriptors for flexible retinal image registration}

\newif\ifuniqueAffiliation
\uniqueAffiliationtrue

\ifuniqueAffiliation 

\usepackage{authblk}

\newbox{\orcid}\sbox{\orcid}{\includegraphics[scale=0.06]{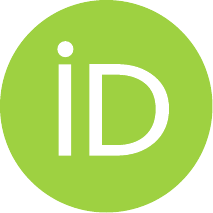}} 
\author[1,2]{%
	\href{https://orcid.org/0000-0001-7824-8098}{\usebox{\orcid}\hspace{1mm}David Rivas-Villar\thanks{\texttt{Corresponding author: david.rivas.villar@udc.es}}}%
}
\author[1,2]{%
	\href{https://orcid.org/0000-0002-9080-9836}{\usebox{\orcid}\hspace{1mm}Álvaro S. Hervella\thanks{\texttt{a.suarezh@udc.es}}}%
}
\author[1,2]{%
	\href{https://orcid.org/0000-0003-4407-9091}{\usebox{\orcid}\hspace{1mm}José Rouco \thanks{\texttt{jrouco@udc.es}}}%
}
\author[1,2]{%
	\href{https://orcid.org/0000-0002-0125-3064}{\usebox{\orcid}\hspace{1mm}Jorge Novo\thanks{\texttt{jnovo@udc.es}}}%
}
\affil[1]{Grupo VARPA, Instituto de Investigacion Biomédica de A Coru\~na (INIBIC), Universidade da Coru\~na, 15006 A Coru\~na, Spain}
\affil[2]{Departamento de Ciencias de la Computación y Tecnologías de la Información, Universidade da Coru\~na, A Coruña, 15071, A Coruña, Spain}
\fi

\renewcommand{\shorttitle}{Unsupervised keypoint-agnostic descriptors for retinal image registration}

\hypersetup{
pdftitle={Unsupervised training of keypoint-agnostic descriptors for flexible retinal image registration},
pdfsubject={cs.CV},
pdfauthor={David~Rivas-Villar, Álvaro S.~Hervella, José~Rouco, Jorge~Novo},
pdfkeywords={Medical image registration, Feature-based Registration, Retinal Image Registration, Medical Imaging},
}

\begin{document}
\maketitle

\begin{abstract}
Current color fundus image registration approaches are limited, among other things, by the lack of labeled data, which is even more significant in the medical domain, motivating the use of unsupervised learning. Therefore, in this work, we develop a novel unsupervised descriptor learning method that does not rely on keypoint detection. This enables the resulting descriptor network to be agnostic to the keypoint detector used during the registration inference.

To validate this approach, we perform an extensive and comprehensive comparison on the reference public retinal image registration dataset. Additionally, we test our method with multiple  keypoint detectors of varied nature, even proposing some novel ones. Our results demonstrate that the proposed approach offers accurate registration, not incurring in any performance loss versus supervised methods. Additionally, it demonstrates accurate performance regardless of the keypoint detector used. Thus, this work represents a notable step towards leveraging unsupervised learning in the medical domain.

\end{abstract}


\section{Introduction}
Image registration is a vital task in  modern healthcare, serving various clinical purposes such as enabling analysis of images from different temporal revisions. However, manual alignment is impractical in the time-constrained clinical setting, necessitating automated registration methods.

In image registration, two images are aligned according to their visual content. The registration pair is composed of a fixed image, which generally remains unchanged, and a moving image, which is transformed to match the contents of the fixed image. Because these pairs capture the same subject under different conditions (e.g., morphology variations, viewpoint, lighting), their data only partially overlaps.

Retinal Image Registration (RIR) is the process of matching and registering images of the retina following their content. Particularly, certain structures are specifically useful to register the images of the human retina. The relevant structures include the blood vessels and optic disk, among others, while the background is usually discarded due to its homogeneity and lack of relevant landmarks.

There are multiple imaging techniques aimed at capturing images of the retina such as color fundus, OCT or fundus angiography. Applications using all of these modalities can take advantage of registration. Among these, Color Fundus (CF) imaging is particularly relevant because it is widespread and extremely cost-effective \cite{costeffec}.

CF registration is challenging due to three distinct factors. First, CF images are prone to multiple imaging defects, including blur, underexposure, overexposure, glares, motion artifacts, etc. Secondly, since CF images capture the retina, they exhibit unique characteristics specific to retinal imaging, such as the particular patterns relevant for registration (e.g., blood vessels), which render general medical image registration methods ineffective. Finally, morphological changes in the retina, such as the appearance or disappearance of blood vessels and the development of lesions (e.g., cotton wool spots), significantly alter the images' appearance and complicate the registration process.

While classical non-deep learning approaches have historically dominated medical image registration, modern deep learning methods now offer comparable or superior results alongside several advantages. Data-driven deep learning eliminates manual tuning, increasing robustness and flexibility over classical methods. However, the severe scarcity of medical labeled data limits traditional, supervised deep learning, motivating the creation of unsupervised methods that require no labeled data.

Generally, current medical image registration methods are not keypoint-based, which makes them less effective for retinal images, particularly CF images. This is evidenced by the fact that the best-performing methods for CF registration are keypoint-based approaches, as any other approach fails to achieve the same performance due to, for instance, ineffective similarity metrics or unrealistic transformations. While deep learning-based registration methods are now predominant in this field, classical methods still achieve the best numerical results, despite their limitations \cite{votus}. The most effective deep learning methods currently involve complex pipelines, utilizing either supervised domain-specific detectors or detector-free techniques \cite{rivas3, eccv20}.

We propose a straight-forward approach for unsupervised descriptor training that allows to use any arbitrary keypoint detector. This eliminates the dependency on labeled data and creates a detector-agnostic keypoint descriptor with all the advantages of deep learning (e.g., flexibility).

We test the unsupervised description network in combination with a series of representative detectors: classical keypoint detectors, anatomical features, and combinations of both, including novel detectors proposed in this work. Based on experiments on the public FIRE dataset, we conclude that our unsupervised descriptor network incurs no performance penalty relative to the supervised baseline. Moreover, we introduce a novel experimental setting to evaluate the trade-off between the number of keypoints and performance. Results show that our unsupervised descriptor is keypoint-agnostic, performing consistently  well across all detectors. By achieving competitive, state-of-the-art results without labeled data, our method overcomes the limitations of prior unsupervised state of the art works. To promote reproducibility and future research, the codebase and trained models will be made publicly available upon publication \footnote{https://github.com/davidrrvv/unconked-retinal-registration}.

\section{Related Work}

Currently, CF registration is evaluated using the public benchmark dataset FIRE \cite{fire}. Notable classical approaches include VOTUS \cite{votus} and REMPE \cite{rempe}. VOTUS matches arteriovenous graphs using optimal transport and classical features, dynamically selecting global parametric transformations (up to low-order quadratic models). REMPE matches both domain-specific and generic keypoints, optimizing correspondences via particle-swarm and RANSAC using a transformation model tailored to fundus images.

In terms of deep learning methods, multiple approaches obtain accurate performance. Current approaches can be divided into detector-based methods and detector-less ones (i.e., with or without keypoint detector). Importantly, unsupervised detector-based methods are, currently, unable to compete with the rest of the approaches in terms of results, despite their advantages. Notably, deep learning methods use homographic transformations.

Among supervised methods, SuperRetina with Knowledge Distillation \cite{kdsr} leverages reverse distillation to train a heavier model, yielding marginal improvements over its predecessor, SuperRetina \cite{eccv20}. 
SuperRetina jointly trains a detector and descriptor using ground-truth vessel keypoints and employs a keypoint-expansion module to detect extra points beyond those initially included in the ground truth.

Addressing SuperRetina’s reliance on on large numbers of keypoints, ConKeD \cite{rivas3} and ConKeD++ \cite{rivas4} achieve equivalent results using significantly fewer keypoints. The increased sparsity reduces computational complexity and enhances robustness. ConKeD utilizes a separate supervised network to extract domain-specific keypoints (blood vessel crossovers and bifurcations) and compute their descriptors. However, a critical limitation of both the SuperRetina and ConKeD families is that their descriptors are conditioned exclusively on their own detectors, making them incompatible with other detectors.

The most relevant detector-less method for CF images is GeoFormer \cite{geoformer}. GeoFormer is based on the pipeline of LoFTR \cite{loftr} which, using transformers, extracts dense matches from the image pair at coarse level and then refines the accurate matches to a finer level later on. GeoFormer \cite{geoformer} adds RANSAC to this pipeline, improving the performance by considering both spatial and image similarity constraints.

Although current deep learning methods trail classical baselines (e.g., VOTUS \cite{votus}) overall, they outperform them in the clinically-relevant Category A of FIRE. To unify the strengths of both paradigms, we propose an unsupervised description network decoupled from the detection phase. This allows our data-driven descriptors to be paired with arbitrary keypoint detectors (classical, learned, or both) enabling highly flexible domain adaptation.

Finally, a new class of keypoint-based methods is emerging that focuses on deformable, non-rigid approaches. These frameworks typically require an initial rigid or homographic registration pass for global alignment, followed by a higher-order deformable transformation to refine the alignment of non-linear structural distortions \cite{global_local, retinaregnet, gauss_primitive}. While these techniques yield impressive local precision, our proposed method explicitly targets the optimization of purely global, rigid registration. Consequently, benchmarking against frameworks that utilize dense, local deformable transformations falls outside the scope of this work.


\section{Methodology}

Our proposal builds upon the description approach of ConKeD++ \cite{rivas4}. However, instead of using a supervised keypoint detector to train the descriptors, we propose to randomly sample keypoints from the retinal fundus. This makes the resulting description network unsupervised and keypoint-agnostic (i.e., applicable to any arbitrary keypoint detector). In that line, we propose to use a set of representative keypoint detectors like anatomical keypoints (including those used in ConKed++), classical keypoints, and combinations of both, alongside our proposed description network.

After the computation of both keypoints and descriptors, the descriptors are matched and then, the paired keypoints, are used to estimate the transformation using RANSAC. It should be noted that, both the detection and description are carried out using an input image size of 565$\times$565, making the obtained pipeline and results equivalent to ConKeD++. Importantly, our description network can be trained with any arbitrary input image size. The transformation estimation through RANSAC is done in the original test set (FIRE) image size, enabling direct comparison with the state of the art.  Like most of the state of the art methods, we use homographic transformations.

\subsection{Unsupervised keypoint training}
\begin{figure}
    \centering
    \includegraphics[width=0.6\textwidth]{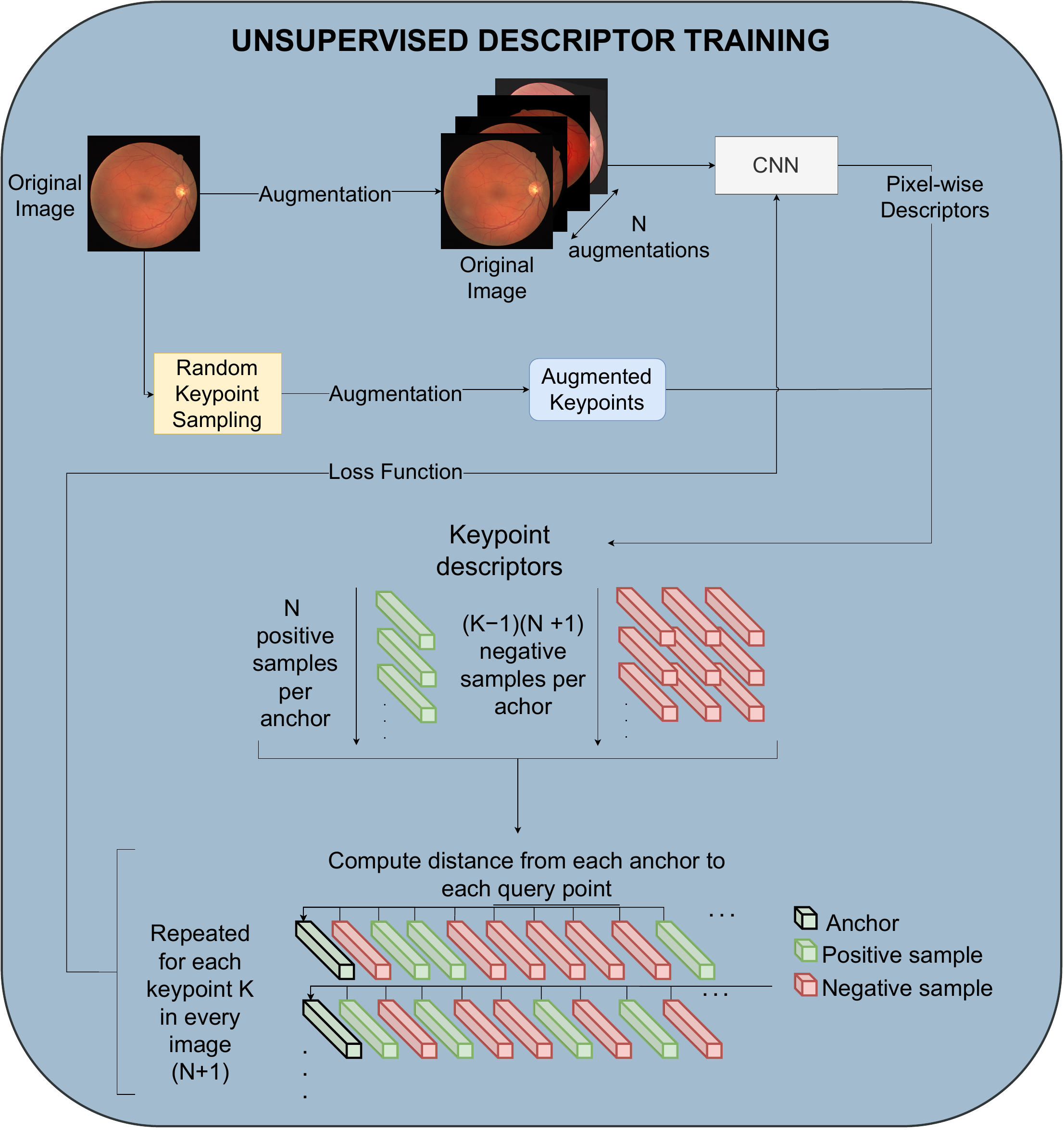}
    \caption{Proposed unsupervised descriptor training}
    \label{fig:method}
\end{figure}

Our proposal for unsupervised descriptor learning builds upon the state of the art work ConKeD++. This method learns descriptors for particular keypoints (blood vessel crossovers and bifurcations) and relies on a supervised detection network to locate them. In this work, we propose UnConKeD (\textbf{Un}supervised \textbf{ConKeD}++), which extends ConKeD++ for unsupervised descriptor learning without a detector, eliminating the dependency on labeled data. Thus, the description network is keypoint-agnostic meaning that it can work with any arbitrary keypoint detector. An overview of the training methodology can be seen in Figure \ref{fig:method}.

We take advantage of the multi-view batch proposed in ConKeD containing multiple views (i.e., augmentations) of the same original image ($1$ original image and $N$ extra views). However, while in the original ConKeD and ConKeD++ the keypoints are sampled from the images following the inference of a supervised detector network (thus dependent on labeled data), in UnConKeD we propose to sample keypoints randomly from the RoI (Region of Interest) of the original image.  This modification serves two primary purposes: firstly, it eliminates the dependency on labeled data, and secondly, it enables the framework to learn descriptors for any arbitrary keypoint on the retinal surface. Typically, replacing supervised keypoints with random sampling imposes a trade-off, eliminating supervision at the cost of a lower signal-to-noise ratio and, possibly, worse learning. However, our methodological design avoids this penalty, successfully decoupling the descriptor without sacrificing performance. Following the keypoint sampling, the network generates dense local descriptor maps associated to each pixel position for each image in the batch. Descriptors corresponding to the sampled keypoints are selected to compute the loss. Following ConKeD++, we use Fast AP loss \cite{fastap,rivas4}.

A key difference between the ConKeD approach and our proposal is the number of sampled keypoints used in training. As ConKeD uses a detector network, the number of keypoints is limited by the vascular structure of the retina. In the proposed approach, the keypoints are freely sampled from the RoI, that is, keypoints are not constrained by morphological structures. Therefore, our approach always samples a fixed amount of keypoints, removing randomness from the training. Theoretically, while random sampling introduces points with a lower signal-to-noise ratio, it provides two critical advantages over structured keypoints. First, structured keypoints restrict the learning signal to specific and sparse morphological regions. Random sampling forces the network to map the entire retinal surface, compelling it to learn generalized and discriminative features even in challenging, texture-sparse areas. Second, extracting this maximized and constant number of keypoints provides denser, more stable and diverse gradients during learning, yielding a more robust and detector-agnostic feature space. Furthermore, as we remove any dependency on supervision, we can use any dataset, with or without labels.

All these improvements speed up the training process as the network receives more and, potentially, better and more diverse feedback in each batch. However, our approach also has some disadvantages. Randomly sampling keypoints in CF images means that many low-interest background points are inevitably sampled (due to the amount of area that is background versus relevant areas). While  a portion of background points may be discernible from others due to proximity to actual landmarks, some of them will be unreliable low-discriminativity points. Although this may slow convergence, improvements in the number of sampled keypoints and dataset flexibility help offset this.

\subsection{Keypoint detection}

We explore diverse keypoint detection strategies by leveraging our unsupervised keypoint-agnostic descriptor network, allowing for a comprehensive comparison of multiple pipeline variations. Specifically, we evaluate:

\textbf{Random Keypoints.} As a baseline, we employ random keypoints, generated by an equispaced grid. Differences in image content ensure that these keypoints do not appear in the same position across registration pairs.

\textbf{Classic keypoint detectors.} These detectors have been used in many applications, even if they are now mostly superseded by deep learning methods. We evaluate the following standard methods: SIFT (Scale-Invariant Feature Transform) \cite{sift}, the Harris corner detector \cite{harris}, FAST (Features from Accelerated Segment Test) \cite{fast}, ORB (Oriented FAST and Rotated BRIEF) \cite{orb}, and CenSurE (Center Surround Extremas) \cite{censure}. For CenSurE, we specifically employ the STAR approximation, which efficiently detects local extrema by approximating Laplacian of Gaussian filters using two overlapping squares.

    \textbf{Blood vessel crossovers and bifurcations.} These domain specific keypoints ensure robust and accurate registration, as demonstrated in ConKeD \cite{rivas3,rivas4}.

    \textbf{Blood vessel segmentations.} We segment the blood vessels in CF images using a state of the art network  \cite{alvaro_asoc}. To ensure full reproducibility of our pipeline, the resulting segmentation masks are published alongside our codebase. Furthermore, our registration framework is completely agnostic to the specific segmentation method used. This modularity allows for the seamless substitution of our masks with those generated by alternative standard models when adapting the pipeline to new datasets. Unlike sparse landmarks, such as crossovers and bifurcations, dense segmentations offer better coverage of the retinal surface, which can improve the registration results. However, this approach remains untested in the SOTA, likely due to the computational complexity of larger keypoint counts. To mitigate this, we propose reducing keypoints via skeletonization or Canny edge detection.

    \textbf{Classic keypoint detectors over blood vessel segmentation logits.} We propose using the logits from the vessel segmentation network as input for classic keypoint detector methods, as this pre-processed data could enhance their performance. This approach remains untested in the SOTA.

\subsection{Training details}

To train the descriptor network we follow the reference ConKeD++ \cite{rivas4} methodology. Thus, we employ the same network and input image size. In particular,  the network is trained from scratch for 1000 epochs with Adam as the optimizer with a fixed learning rate of $1e-4$. The number of bins used in the FastAP Loss, was set to $Q=10$. To train, we use the Messidor-2 dataset \cite{mess,mess2}, normalizing image  size to $565\times565$ (ConKeD++ input size). We also use the exact same augmentation regime used in ConKeD++. We use random affine transformations for spatial augmentations. These transformations include rotations of $\pm 60^{\circ}$, translations of $0.25\times imageSize$ in each axis, scaling between $0.75-1.25 \times imageSize$ and shearing of $\pm 30^{\circ}$. For color augmentation, we use random changes in the HSV color space and Gaussian noise (mean 0, standard deviation 0.05), with noise applied 25\% of the time. Both spatial and HSV augmentations are applied to every augmented image in the batch. Following ConKeD++, each batch contains one original image and nine augmentations (batch size 10). Leveraging the controlled sampling process, we select the hardware maximum of 1460 keypoints per image, conducting training on two 80GB Nvidia A100 GPUs.
Qualitative observations indicate that maximizing this keypoint count accelerates convergence. Furthermore, spatial augmentations proved strictly necessary to achieve geometric invariance, while color and noise-based augmentations prevent overfitting to specific lighting conditions or device specific settings.

\subsection{Matching details}

Improving efficiency in detector-based pipelines requires reducing keypoints to decrease comparisons during matching and combinations during transformation estimation. Therefore, finding detectors with the best performance-per-keypoint ratio is highly desirable. To ensure a fair comparison, detectors are evaluated at predefined keypoint targets. We tune the different sensitivity parameters of each detector so that the average number of keypoints detected across the entire test set matches our targets. We evaluate performance at average counts of 100, 500, and 1000 keypoints, as well as an \textit{unlimited} setting where all thresholds are removed. For random keypoints, we add extra levels of 5000 and 40000 keypoints to account for their expected lower performance.

The blood vessel crossovers and bifurcations detector herein tested is the same used in ConKeD++ \cite{rivas4}, making the results directly comparable. 

\subsection{Datasets and evaluation}

To train the unsupervised description network we use the Messidor-2 dataset \cite{mess,mess2}, which consits of 1748 images captured in a set of Diabetic Retinopathy examinations. The images are captured using a 45º of FOV. For evaluation, we use the FIRE dataset, which is the only one with suitable ground truth for CF registration. It contains 129 retinal images from 39 patients, generating 134 registration pairs. The dataset is divided into three categories: Category S (71 pairs) with high overlap, Category P (49 pairs) with low overlap, and Category A (14 pairs) with high overlap and disease progression, making it particularly challenging and highly clinically relevant. 
We evaluate the registration using the FIRE proposed metric, Registration Score, at the dataset's original resolution \cite{fire}. Registration Score is based on the Euclidean distance among the the ground truth points. Plotting this distance against a mobile error threshold allows us to calculate an Area Under the Curve (AUC) metric, which is the final Registration Score.

\section{Results and Discussion}

\subsection{Supervised vs Unsupervised descriptor training}

Table \ref{tab:dsc} compares the performance of the supervised and unsupervised descriptor networks. Because crossovers and bifurcations yield few keypoints, we can eliminate RANSAC randomness entirely by running the algorithm without a maximum iteration budget. This exhaustive approach guarantees the selection of the optimal transformation for every image pair. Notably, compared to ConKeD++, the unsupervised training achieves a nominally higher overall performance. It is worth noting that this comparison uses vessel crossovers and bifurcations as keypoints, and ConKeD++ has been specifically trained to produce descriptors for these keypoints, meaning it is the most favorable setup for ConKeD++. In this setting, even performance parity would be an achievement. To rigorously evaluate the difference between the methods, we performed a Wilcoxon signed-rank test across the dataset pairs using these deterministic scores. The test yielded no statistically significant difference ($p = 0.42$), confirming that the methods perform equally well. This represents a meaningful step forward in unsupervised learning as we successfully match supervised performance without requiring labeled training data, while even securing a slight empirical edge in the aggregated metrics.

The proposed random keypoint sampling strategy also has advantages at inference time, beyond extra flexibility resulting from forgoing labeled data. In ConKeD++, the lack of feedback for non-crossover or bifurcation points can lead to inaccurate descriptors and incorrect matching, if any point other that those is mistakenly detected. In contrast, our approach trains with points from the entire retinal surface, enabling accurate descriptors for all retinal areas, improving its robustness. In summary, the unsupervised process adds no drawbacks while providing significant benefits, particularly in the medical domain where labels are often scarce.

\begin{table}[tbp]
\centering
\resizebox{0.55\textwidth}{!}{%
\begin{tabular}{@{}llccccc@{}}
\toprule
FIRE Dataset                                                      &  FIRE   & A      & P       & S     & Avg.   & W. Avg. \\ \midrule
CB + ConKeD++ \cite{rivas4} & 0.765   & 0.766  & 0.503   & 0.945 & 0.738  & 0.765   \\
   CB + UnConKeD    & 0.774  & 0.760  & 0.527   & 0.948 & 0.745  & 0.774   \\  \bottomrule
\end{tabular}%
}
\vspace{1em}
\caption{Comparison between the supervised and unsupervised descriptor training using crossovers and bifurcations as the keypoints}
\label{tab:dsc}
\end{table}

\subsection{Keypoint detectors}

\begin{figure*}
    \centering
    \includegraphics[width=.95\textwidth]{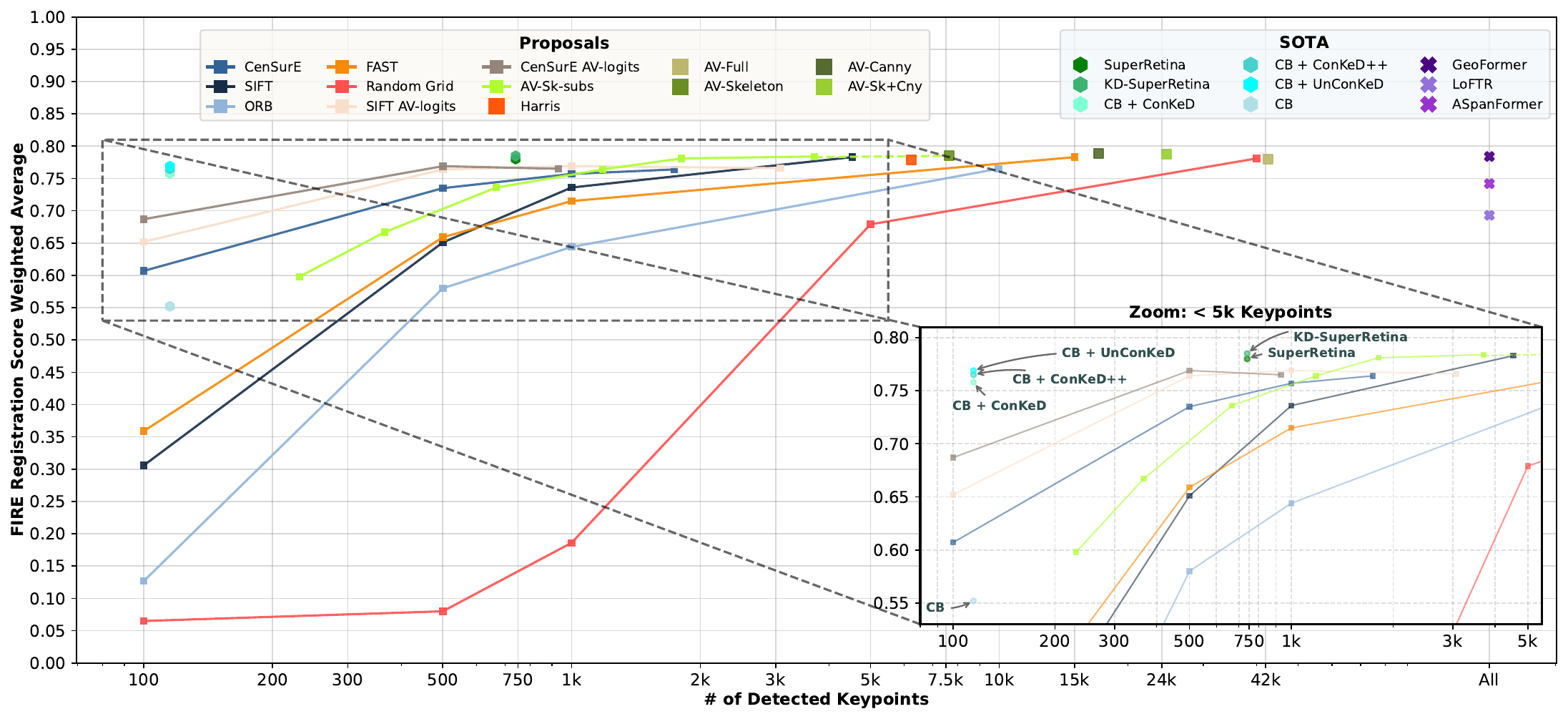}
    \caption{Results for the different keypoint detectors in FIRE, measured in Weighted Average of Registration Score AUC}
    \label{fig:res}
\end{figure*}

The results of combining our unsupervised keypoint descriptor with different keypoint detectors are shown in Table \ref{tab:res}. Moreover, the results are shown in a intuitive graphical visualization in Figure \ref{fig:res}.

\begin{table}[tbp]
\centering
\resizebox{0.95\textwidth}{!}{%
\begin{tabular}{@{}llcccccc@{}}
\toprule
FIRE Dataset & \multicolumn{1}{c}{\begin{tabular}[c]{@{}c@{}}Average \# \\ of keypoints\end{tabular}} & FIRE & A & P & S & Avg. & W. Avg. \\ \midrule
CB$\dagger$ & 115 & 0.774  & 0.760  & 0.527   & 0.948 & 0.745  & 0.774 \\ \midrule

\multirow{5}{*}{\begin{tabular}[c]{@{}l@{}}Random Keypoint\\ Grid\end{tabular}} 
 & 100 & 0.065 {\small $\pm$ 0.000} & 0.054 {\small $\pm$ 0.000} & 0.000 {\small $\pm$ 0.000} & 0.113 {\small $\pm$ 0.000} & 0.056 {\small $\pm$ 0.000} & 0.065 {\small $\pm$ 0.000} \\
 & 500 & 0.079 {\small $\pm$ 0.002} & 0.074 {\small $\pm$ 0.007} & 0.006 {\small $\pm$ 0.001} & 0.130 {\small $\pm$ 0.002} & 0.070 {\small $\pm$ 0.003} & 0.079 {\small $\pm$ 0.002}\\
 & 1,000 & 0.183 {\small $\pm$ 0.009} & 0.209 {\small $\pm$ 0.023} & 0.091 {\small $\pm$ 0.006} & 0.241 {\small $\pm$ 0.008} & 0.180 {\small $\pm$ 0.012} & 0.183 {\small $\pm$ 0.009} \\
 
 & 5,000 & 0.684 {\small $\pm$ 0.006} & 0.738 {\small $\pm$ 0.008} & 0.467 {\small $\pm$ 0.014} & 0.823 {\small $\pm$ 0.002} & 0.676 {\small $\pm$ 0.005} & 0.684 {\small $\pm$ 0.006}\\

& 40,000 & \textit{0.781 {\small $\pm$ 0.001}} & \textit{0.774 {\small $\pm$ 0.008}} & \textit{0.542 {\small $\pm$ 0.003}} & \textit{0.947 {\small $\pm$ 0.000}} & \textit{0.754 {\small $\pm$ 0.003}} & \textit{0.781 {\small $\pm$ 0.001}} \\ \midrule
 
\multirow{4}{*}{CENSURE} & 
100 & 0.607 {\small $\pm$ 0.002} & 0.527 {\small $\pm$ 0.008} & 0.268 {\small $\pm$ 0.006} & 0.856 {\small $\pm$ 0.006} & 0.550 {\small $\pm$ 0.002} & 0.607 {\small $\pm$ 0.002}
 \\

 & 500 & 0.735 {\small $\pm$ 0.003} & 0.721 {\small $\pm$ 0.015} & 0.464 {\small $\pm$ 0.005} & 0.926 {\small $\pm$ 0.001} & 0.703 {\small $\pm$ 0.006} & 0.735 {\small $\pm$ 0.003} \\
 
 & 1,000 & 0.757 {\small $\pm$ 0.002} & 0.738 {\small $\pm$ 0.006} & 0.496 {\small $\pm$ 0.004} & \textit{0.941 {\small $\pm$ 0.002}} & 0.725 {\small $\pm$ 0.003} & 0.757 {\small $\pm$ 0.002} \\
 
& 1,740 (all) & \textit{0.764 {\small $\pm$ 0.003}} & \textit{0.754 {\small $\pm$ 0.015}} & \textit{0.514 {\small $\pm$ 0.006}} & 0.939 {\small $\pm$ 0.001} & \textit{0.736 {\small $\pm$ 0.006}} & \textit{0.764 {\small $\pm$ 0.003}} \\
 \midrule

\multirow{4}{*}{ORB} 
 & 100 & 0.127 {\small $\pm$ 0.005} & 0.189 {\small $\pm$ 0.012} & 0.049 {\small $\pm$ 0.006} & 0.168 {\small $\pm$ 0.008} &  {0.135 \small $\pm$ 0.006} & 0.127 {\small $\pm$ 0.005} \\
 
 & 500 & 0.580 {\small $\pm$ 0.005} & 0.445 {\small $\pm$ 0.030} & 0.207 {\small $\pm$ 0.002} & 0.865 {\small $\pm$ 0.005}& 0.505 {\small $\pm$ 0.011} & 0.580 {\small $\pm$ 0.005} \\
 & 1,000 & 0.644 {\small $\pm$ 0.007} & 0.532 {\small $\pm$ 0.056} & 0.287 {\small $\pm$ 0.006} & 0.913 {\small $\pm$ 0.002} & 0.577 {\small $\pm$ 0.019} & 0.644 {\small $\pm$ 0.007} \\
 
& 9,953 (all) & \textit{0.765 {\small $\pm$ 0.003}} & \textit{0.771 {\small $\pm$ 0.010}} & \textit{0.492 {\small $\pm$ 0.007}} & \textit{0.952 {\small $\pm$ 0.001}} & \textit{0.739 {\small $\pm$ 0.004}} & \textit{0.765 {\small $\pm$ 0.003}}\\ 
 
 \midrule
\multirow{4}{*}{SIFT} 
 & 100 & 0.306 {\small $\pm$ 0.012} & 0.272 {\small $\pm$ 0.013} & 0.140 {\small $\pm$ 0.006} & 0.427 {\small $\pm$ 0.018} & 0.280 {\small $\pm$ 0.011} & 0.306 {\small $\pm$ 0.012} \\
 
 & 500 & 0.651 {\small $\pm$ 0.004} & 0.510 {\small $\pm$ 0.030} & 0.347 {\small $\pm$ 0.004}  & 0.889 {\small $\pm$ 0.002} & 0.582 {\small $\pm$ 0.010} & 0.651 {\small $\pm$ 0.004} \\
 
 & 1,000 & 0.736 {\small $\pm$ 0.003} & 0.741 {\small $\pm$ 0.014} & 0.442 {\small $\pm$ 0.008} & 0.937 {\small $\pm$ 0.001} & 0.707 {\small $\pm$ 0.004} & 0.736 {\small $\pm$ 0.003} \\
 
& 4,529 (all) & \textit{0.783 {\small $\pm$ 0.002}} & \textit{0.774 {\small $\pm$ 0.009}} & \textit{0.537 {\small $\pm$ 0.004}} & \textit{0.954 {\small $\pm$ 0.001}} & \textit{0.755 {\small $\pm$ 0.003}} & \textit{0.783 {\small $\pm$ 0.002}}\\ 
 \midrule
 
\multirow{4}{*}{FAST} 
 & 100 & 0.359 {\small $\pm$ 0.009} & 0.386 {\small $\pm$ 0.015} & 0.138 {\small $\pm$ 0.005} & 0.887 {\small $\pm$ 0.001} & 0.343 {\small $\pm$ 0.003} & 0.359 {\small $\pm$ 0.009} \\
 
 & 500 & 0.659 {\small $\pm$ 0.004} & 0.610 {\small $\pm$ 0.027}& 0.345 {\small $\pm$ 0.005} & 0.887 {\small $\pm$ 0.001} & 0.614 {\small $\pm$ 0.010} & 0.659 {\small $\pm$ 0.004} \\
 
 & 1,000 & 0.715 {\small $\pm$ 0.003} & 0.714 {\small $\pm$ 0.027} & 0.409 {\small $\pm$ 0.006} & 0.926 {\small $\pm$ 0.001} & 0.683 {\small $\pm$ 0.009} & 0.715 {\small $\pm$ 0.003} \\
 
& 15,020 (all) & \textit{0.783 {\small $\pm$ 0.001}} & \textit{0.768 {\small $\pm$ 0.005}} & \textit{0.543 {\small $\pm$ 0.001}} & \textit{0.952 {\small $\pm$ 0.001}} & \textit{0.755 {\small $\pm$ 0.001}} & \textit{0.783 {\small $\pm$ 0.001}} \\ 
 \midrule

Harris & 6,224 & 0.779 {\small $\pm$ 0.001} & 0.760 {\small $\pm$ 0.006} & 0.537 {\small $\pm$ 0.005} & 0.950 {\small $\pm$ 0.001} & 0.749 {\small $\pm$ 0.001} & 0.779 {\small $\pm$ 0.001} \\ \midrule

AV All Points & 42,518 & 0.780 {\small $\pm$ 0.001} & 0.780 {\small $\pm$ 0.003} & 0.522 {\small $\pm$ 0.004} & \textit{\textbf{0.958 {\small $\pm$ 0.000}}} & 0.754 {\small $\pm$ 0.002} & 0.780 {\small $\pm$ 0.001}\\

AV Skeleton & 7,648 & 0.785 {\small $\pm$ 0.001} & 0.784 {\small $\pm$ 0.007} & 0.538 {\small $\pm$ 0.003} & 0.956 {\small $\pm$ 0.000} & 0.759 {\small $\pm$ 0.002} & 0.785 {\small $\pm$ 0.001} \\

AV Canny & 17,054 & \textit{\textbf{0.789 {\small $\pm$ 0.001}}} & 0.783 {\small $\pm$ 0.002} & \textit{\textbf{0.551 {\small $\pm$ 0.003}}} & 0.956 {\small $\pm$ 0.001} & \textit{\textbf{0.763 {\small $\pm$ 0.001}}} & \textit{\textbf{0.789 {\small $\pm$ 0.001}}} \\

AV Skelet+Canny & 24,631 & 0.788 {\small $\pm$ 0.002} & \textit{\textbf{0.784 {\small $\pm$ 0.003}}} & 0.546 {\small $\pm$ 0.005} & 0.956 {\small $\pm$ 0.000} & 0.762 {\small $\pm$ 0.003} & 0.788 {\small $\pm$ 0.002}\\ 

\midrule
\multirow{4}{*}{\begin{tabular}[c]{@{}l@{}}SIFT over\\ AV logits\end{tabular}} 

 & 100 & 0.652 {\small $\pm$ 0.005} & 0.567 {\small $\pm$ 0.010} & 0.355 {\small $\pm$ 0.008} & 0.874 {\small $\pm$ 0.004} & 0.599 {\small $\pm$ 0.005} & 0.652 {\small $\pm$ 0.005} \\
 
 & 500 & 0.764 {\small $\pm$ 0.002} & 0.759 {\small $\pm$ 0.004} & 0.505 {\small $\pm$ 0.007} & 0.944 {\small $\pm$ 0.001} & 0.736 {\small $\pm$ 0.002} & 0.764 {\small $\pm$ 0.002} \\
 
& 1,000 & \textit{0.769 {\small $\pm$ 0.002}} & \textit{0.762 {\small $\pm$ 0.009}} & \textit{0.518 {\small $\pm$ 0.004}} & \textit{0.944 {\small $\pm$ 0.001}} & \textit{0.741 {\small $\pm$ 0.003}} & \textit{0.769 {\small $\pm$ 0.002}} \\
 
 & 3,079 (all) & 0.766 {\small $\pm$ 0.004} & 0.757 {\small $\pm$ 0.010} & 0.509 {\small $\pm$ 0.010} & 0.945 {\small $\pm$ 0.001} & 0.737 {\small $\pm$ 0.004} & 0.766 {\small $\pm$ 0.004} \\ \midrule
\multirow{3}{*}{\begin{tabular}[c]{@{}l@{}}CENSURE over\\ AV logits\end{tabular}} 
 & 100 & 0.687 {\small $\pm$ 0.004} & 0.618 {\small $\pm$ 0.016} & 0.420 {\small $\pm$ 0.007} & 0.886 {\small $\pm$ 0.001} & 0.641 {\small $\pm$ 0.007} & 0.687 {\small $\pm$ 0.004}\\
 
& 500 & \textit{0.769 {\small $\pm$ 0.002}} & \textit{0.740 {\small $\pm$ 0.006}} & \textit{0.522 {\small $\pm$ 0.006}} & \textit{0.945 {\small $\pm$ 0.000}} & \textit{0.736 {\small $\pm$ 0.002}} & \textit{0.769 {\small $\pm$ 0.002}}
 \\
 & 932 (all) & 0.765 {\small $\pm$ 0.004} & 0.731 {\small $\pm$ 0.018} & 0.516 {\small $\pm$ 0.012} & 0.944 {\small $\pm$ 0.001} & 0.730 {\small $\pm$ 0.006} & 0.765 {\small $\pm$ 0.004} \\ \midrule
\multirow{6}{*}{\begin{tabular}[c]{@{}l@{}}AV Skeleton\\ Random Subsampling\end{tabular}} 

 & 231 & 0.598 {\small $\pm$ 0.004}  & 0.569 {\small $\pm$ 0.015}  & 0.361 {\small $\pm$ 0.011}  & 0.767 {\small $\pm$ 0.006}  & 0.566 {\small $\pm$ 0.005} & 0.598 {\small $\pm$ 0.004}  \\

  & 366 & 0.667 {\small $\pm$ 0.002} & 0.674 {\small $\pm$ 0.024} & 0.427 {\small $\pm$ 0.005} & 0.831 {\small $\pm$ 0.003} & 0.644 {\small $\pm$ 0.008} & 0.667 {\small $\pm$ 0.002} \\

   & 667 & 0.736 {\small $\pm$ 0.002} & 0.754 {\small $\pm$ 0.004} & 0.489 {\small $\pm$ 0.006} & 0.902 {\small $\pm$ 0.002} & 0.715 {\small $\pm$ 0.002} & 0.736 {\small $\pm$ 0.002} \\

  & 1,182 & 0.764 {\small $\pm$ 0.001} & 0.773 {\small $\pm$ 0.006} & 0.507 {\small $\pm$ 0.003} & 0.940 {\small $\pm$ 0.001} & 0.740 {\small $\pm$ 0.002} & 0.764 {\small $\pm$ 0.001} \\

   & 1,807 & 0.781 {\small $\pm$ 0.003} & 0.777 {\small $\pm$ 0.004} & 0.533 {\small $\pm$ 0.007} & 0.953 {\small $\pm$ 0.000} & 0.754 {\small $\pm$ 0.003}& 0.781 {\small $\pm$ 0.003} \\
 
 & 3,696 & \textit{0.784 {\small $\pm$ 0.002}} & \textit{0.781 {\small $\pm$ 0.007}} & \textit{0.534 {\small $\pm$ 0.005}} & \textit{0.957 {\small $\pm$ 0.000}} & \textit{0.757 {\small $\pm$ 0.003}} & \textit{0.784 {\small $\pm$ 0.002}} \\

 \bottomrule
\end{tabular}%
}

\vspace{1em}
\caption{Results in FIRE for the different detectors combined with the unsupervised descriptors, measured in Registration Score AUC $\pm$ unbiased sample standard deviation (i.e., with Bessel's correction), derived from 5 independent matching executions (RANSAC). $\dagger$ indicates deterministic results. Best overall results highlighted in \textbf{bold}, best in-block results highlighted in \textit{italics}.}
\label{tab:res}
\end{table}

First, the random keypoint grid approach is ineffective with low keypoint counts. It only performs competitively with 40,000 keypoints, which is impractical due to high matching costs. However, we include it as a baseline for comparison with state-of-the-art methods because it is fully unsupervised as the keypoints are selected randomly without any supervision or extra data.

The classical keypoint detectors offer accurate performance, specially at higher numbers of keypoints. ORB and SIFT require at least 500 keypoints to obtain accurate results. However, CenSurE surpasses 0.6 of AUC at just 100 keypoints. Furthermore, with 500 or 1000 keypoints CenSurE reaches similar levels to the supervised crossovers and bifurcations. Utilizing more keypoints (e.g., all possible) incurs in diminishing returns, as the results do not improve enough to justify the increase in keypoints used. Conversely, since ORB or SIFT exhibit lower performance with fewer keypoints, increasing the number of keypoints becomes more relevant in these methods. SIFT improves the results of CenSurE when using all detectable keypoints. However, this is simply due to SIFT detecting many more keypoints (2.5$\times$). This can also be intuitively seen in Figure \ref{fig:res}.

In terms of corner detectors, FAST performs similar to the keypoint detectors, positioning itself behind SIFT but ahead of ORB. FAST needs at least 500 keypoints to offer useful performance. Removing the sensitivity threshold, it detects around 15k corners, an impractically large number, which allows this method to compete with SIFT and CenSurE.
In contrast, the Harris corner detector offers approximately the same performance but using half of the keypoints, around 6k.

The different alternatives using keypoints derived from the arteriovenous tree are among the best-performing methods. Using all the keypoints in the vessel segmentations is not practical despite the good performance, as it results in more than 40k keypoints. The different strategies for reducing the amount of keypoints proved to be useful and reliable. Both using the Canny edge detector and morphological skeletonization massively reduce the amount of keypoints while keeping or even improving the performance. Regarding the comparison between both approaches, Canny provides double the number of keypoints than the skeleton for minuscule performance gains.  Combining the Canny edges with the skeletonization marginally improves performance compared to using each method separately, while significantly increasing the keypoints. Thus, using the skeleton provides the best balance between performance and computational cost, as it uses around 7.5k keypoints. However, it is still possible to further improve the efficiency of this method, since, due to the structure of the blood vessels, many keypoints are redundant in the transformation estimation. Thus, we  randomly sub-sample these keypoints. By iterating over the skeleton, we remove the closest neighbors of each point using increasingly big kernel sizes. The results of this approach can be seen in Table \ref{tab:res} under the name "AV Skeleton Sub-sampling". Using less than 25\% of the original points (i.e. 1807 as opposed to 7648) we retain the performance while significantly reducing the computational cost. Therefore, this approach can be considered a good compromise between performance and cost, as it offers the best results despite using a limited number of keypoints.

Finally, applying keypoint detectors to the logits of the segmentation network improves performance compared to using the detectors directly on the images. However, this improvement is only relative the number of used keypoints, meaning it improves the efficiency of the method allowing better results with less keypoints. Using the logits as input causes CenSurE and SIFT detect less keypoints, which limits their top performance but the performance-per-keypoint improves.  This can be easily seen in Figure \ref{fig:res} as the results of using logits as input for the detectors are more to the top and left than the detectors with images as input. 

Overall, these results highlight the keypoint-agnostic nature of our unsupervised description network, as it is able to consistently produce accurate registrations with a wide variety of keypoint detectors. Importantly, regarding computational efficiency, the overall runtime of our framework is dictated by the matching phase of the chosen detector, as descriptor inference takes only a fraction of a second. The keypoint-agnostic design uniquely allows users to scale computational costs to their specific hardware limits by simply selecting sparser detectors.

\subsection{State of the Art Comparison}

In this subsection, we compare our approaches with methods from the state of the art. Direct comparisons can be seen in Table \ref{tab:sota} as well as Figure \ref{fig:res}.

\begin{table}[t]
\centering
\resizebox{0.95\textwidth}{!}{%
\begin{tabular}{@{}lcccccc@{}}
\toprule
FIRE Dataset                                   & \multicolumn{1}{r}{FIRE}           & A                                           & P                                  & S                                  & Avg.                                        & W. Avg.                            \\ \midrule
Classical \\

\multicolumn{1}{r}{VOTUS \cite{votus}}            & \underline{\textbf{0.812}}                     & \textbf{0.681}                                       & \underline{\textbf{0.672}}                     & 0.934                              & \textbf{0.762}                                       & \underline{\textbf{0.811}}                     \\

\multicolumn{1}{r}{REMPE \cite{rempe}}           & 0.773                              & 0.66                                        & 0.542                              & \underline{\textbf{0.958}}                     & 0.72                                        & 0.774                              \\

\multicolumn{1}{r}{Harris-PIIFD \cite{harris_piifd, rempe}}	&0.553	&0.443	&0.09	&0.9	&0.478	&0.556 \\
\multicolumn{1}{r}{SURF+WGTM \cite{wang_surf, rempe}}	&0.472	&0.069	&0.061	&0.835	&0.322	&0.472
\\
\\
Deep Learning \\
\multicolumn{1}{r}{\textit{AV Canny}} & \multicolumn{1}{l}{\textbf{\textit{0.789 {\small $\pm$ 0.001}}}} & \multicolumn{1}{l}{\textit{\underline{\textbf{0.783 {\small $\pm$ 0.002}}}}} & \multicolumn{1}{l}{\textit{0.551 {\small $\pm$ 0.003}}} & \multicolumn{1}{l}{\textbf{\textit{0.956 {\small $\pm$ 0.001}}}} & \multicolumn{1}{l}{\underline{\textit{\textbf{0.763 {\small $\pm$ 0.001}}}}} & \multicolumn{1}{l}{\textbf{\textit{0.789 {\small $\pm$ 0.001}}}} \\\\

\multicolumn{1}{r}{KD-SuperRetina \cite{kdsr}}    & -                                  & 0.783                                       & 0.558*                             & 0.942                              & 0.761*                                      & 0.785*                             \\
\multicolumn{1}{r}{GeoFormer  \cite{geoformer}}   & -                                  & 0.760                                        & \textbf{0.559*}                             & 0.944                              & 0.754*                                      & 0.784*                             \\
\multicolumn{1}{r}{\textit{AV Skeleton SubSampled}}                           & \textit{0.781 {\small $\pm$ 0.003}}                     
& \textit{0.777 {\small $\pm$ 0.004}}                              & \textit{0.533 {\small $\pm$ 0.007}}                     & \textit{0.953 {\small $\pm$ 0.000}}                     & \textit{0.754 {\small $\pm$ 0.003}}                              & \textit{0.781 {\small $\pm$ 0.003}}                     \\
\multicolumn{1}{r}{SuperRetina \cite{eccv20}}     & -                                  & 0.783                                       & 0.542*                             & 0.94                               & 0.755*                                      & 0.780*                             \\

\multicolumn{1}{r}{\textit{CenSurE AV Logits 500}}  &

\textit{0.769 {\small $\pm$ 0.002}} & \textit{0.740 {\small $\pm$ 0.006}} & \textit{0.522 {\small $\pm$ 0.006}} & \textit{0.945 {\small $\pm$ 0.000}} & \textit{0.736 {\small $\pm$ 0.002}} & \textit{0.769 {\small $\pm$ 0.002}} \\

\multicolumn{1}{r}{ConKeD++ \cite{rivas4}}        & 0.760                               & 0.766                                       & 0.503                              & 0.945                              & 0.738                                       & 0.765                              \\
\multicolumn{1}{r}{ConKeD \cite{rivas3}}          & 0.758                              & 0.749                                       & 0.489                              & 0.945                              & 0.728                                       & 0.758                              \\
\multicolumn{1}{r}{ASpanFormer  \cite{geoformer}} & -                                  & 0.703                                       & 0.495*                             & 0.921                              & 0.706*                                      & 0.742*                             \\
\multicolumn{1}{r}{LoFTR  \cite{geoformer}}       & -                                  & 0.711                                       & 0.359*                             & 0.92                               & 0.663*                                      & 0.693*                             \\
\multicolumn{1}{r}{Retina-R2D2 \cite{rivas2}}     & 0.695                              & 0.726                                       & 0.352                              & 0.925                              & 0.645                                       & 0.575                              \\

 \bottomrule
\end{tabular}%
}
\vspace{1em}
\caption{Comparison of our proposals (italics) with state-of-the-art methods, sorted by FIRE weighted average. Best results for each category in bold, best overall underlined. * indicates evaluation without the full set of images.}
\label{tab:sota}
\end{table}

Our unsupervised descriptor training, combined with vessel points, surpasses all other state-of-the-art deep learning methods. Among the evaluated approaches, the Canny method yields the best overall performance across all categories except Category P, where GeoFormer scores highest. However, it should be noted that GeoFormer evaluates fewer images, making those specific results not fully comparable to our own. Crucially, because the overall aggregated metric for the SOTA methods falls strictly below our mean-minus-standard-deviation threshold for this method, we can confidently affirm the competitive superiority of our approach. Furthermore, the sub-sampled skeleton method remains highly competitive despite utilizing significantly fewer keypoints. Most importantly, even when accounting for the standard deviation, both the Canny version and the sub-sampled skeleton approach achieve results among the best in Category A, which holds the greatest clinical significance due to its pathological progression. 

While CenSurE with AV Logits ranks slightly lower, its performance remains remarkable as it utilizes only 500 keypoints. It uses fewer keypoints than any other method outperforming it. 

Another interesting comparison to make is between detector-less methods and the random keypoint approach. Detector-less methods such as GeoFormer \cite{geoformer} can be, effectively, unsupervised. However, they use the whole image to obtain matches which is, potentially, more computationally intensive. A direct comparison for these methods would be the random keypoint grid which is fully unsupervised as the keypoints are not generated based on any image feature or learning process. In this regard, our random grid approach using 40k points obtains a weighted registration score of 0.782, which is comparable to the 0.784 from Geoformer. For reference, there are a approximately 215k points in the RoI of the FIRE fundus images at the operating resolution of our description network. Furthermore, Geoformer operates at a higher resolution meaning that it can create finer matches but at even higher computational cost.

Finally, the variants of our method using classical keypoint detectors considerably outperform previous methods using the same type of detectors \cite{harris_piifd, wang_surf}, (e.g., our Harris vs Harris+PIIFD \cite{harris_piifd} improves results over 20\%). Thus, we can conclude that our unsupervised descriptor is able to significantly boost the performance of classical detectors.

\section{Conclusion}

In this work, we present a novel unsupervised method for training keypoint descriptors. It addresses one of the main limiting factors of current learning approaches, the requirement for labeled data, which is specially scarce in the medical domain. Using  random keypoint sampling from the retinal RoI, we create an unsupervised description network that is agnostic to keypoint detectors. In this sense, we test our approach with multiple representative keypoint detectors, including novel ones proposed in this work.

We empirically show that our unsupervised descriptor not only avoids a performance decrease compared to supervised training, but actually improves upon it. This eliminates the typical performance penalty of self-supervised or unsupervised methods. Furthermore, its consistently accurate performance across various keypoint detectors highlights its flexibility and keypoint-agnostic nature. Notably, when combined with several of our proposed keypoint detectors, our pipeline achieves competitive performance with current SOTA, even surpassing it. Our work represents a significant advancement that enables unsupervised methods to perform on par with or surpass supervised approaches, a crucial feat in medical fields. Future work will focus on validating this keypoint-agnostic framework across additional emerging datasets and varying retinal imaging modalities.

\section*{Acknowledgements}

This work was supported by the Instituto de Salud Carlos III (\mbox{ISCIII}), Government of Spain [grant number \mbox{FORT23/00010} (Programa \mbox{FORTALECE})]; \mbox{MICIU}/\mbox{AEI}/\mbox{10.13039/501100011033} and ``\mbox{ERDF} A way of making Europe'' [grant numbers \mbox{PID2023-148913OB-I00} and \mbox{PID2024-161024OB-I00}]; and the Consellería de Educación, Universidade e Formación Profesional, Xunta de Galicia, Grupos de Referencia Competitiva [grant number \mbox{ED431C 2024/33}], pre-doctoral grant number \mbox{ED481A 2021/147}, and post-doctoral fellowship number \mbox{ED481B-2022-025}. Funding for open access charge: Universidade da Coruña/CISUG.

\bibliographystyle{elsarticle-num} 

\bibliography{references}

@ARTICLE{rempe,
  author={C. {Hernandez-Matas} and X. {Zabulis} and A. A. {Argyros}},
  journal={IEEE Journal of Biomedical and Health Informatics}, 
  title={REMPE: Registration of Retinal Images Through Eye Modelling and Pose Estimation}, 
  year={2020},
  volume={24},
  pages={3362-3373},

}

@ARTICLE{harris_piifd,
  author={J. {Chen} and J. {Tian} and N. {Lee} and J. {Zheng} and R. T. {Smith} and A. F. {Laine}},
  journal={IEEE Transactions on Biomedical Engineering}, 
  title={A Partial Intensity Invariant Feature Descriptor for Multimodal Retinal Image Registration}, 
  year={2010},
  volume={57},
  number={7},
  pages={1707-1718},
 }

@article{wang_surf,
title = {Robust point matching method for multimodal retinal image registration},
author = {Gang Wang and Zhicheng Wang and Yufei Chen and Weidong Zhao},
journal = {Biomedical Signal Processing and Control},
volume = {19},
pages = {68-76},
year = {2015},
issn = {1746-8094},
}

@article{fire,
author = {Hernandez-Matas, Carlos and Zabulis, Xenophon and Triantafyllou, Areti and Anyfanti, Panagiota and Douma, Stella and Argyros, Antonis},
pages = {},
title = {FIRE: Fundus Image Registration Dataset},
journal = {Journal for Modeling in Opthalmology}
}

@ARTICLE{votus,
  author={D. {Motta} and W. {Casaca} and A. {Paiva}},
  journal={IEEE Transactions on Image Processing}, 
  title={Vessel Optimal Transport for Automated Alignment of Retinal Fundus Images}, 
  year={2019},
  volume={28},
  number={12},
  pages={6154-6168},
 }

@article{sift,
author={Lowe, David G.},
title={Distinctive Image Features from Scale-Invariant Keypoints},
journal={International Journal of Computer Vision},
year={2004},
month={Nov},
day={01},
volume={60},
pages={91-110},
issn={1573-1405},
}

@inproceedings{orb,
author = {Rublee, Ethan and Rabaud, Vincent and Konolige, Kurt and Bradski, Gary},
year = {2011},
month = {11},
pages = {2564-2571},
title = {ORB: an efficient alternative to SIFT or SURF},
journal = {Proceedings of the IEEE International Conference on Computer Vision},
}

@article{mess,
    author = {Abràmoff, Michael D. and Folk, James C. and Han, Dennis P. and Walker, Jonathan D. and Williams, David F. and Russell, Stephen R. and Massin, Pascale and Cochener, Beatrice and Gain, Philippe and Tang, Li and Lamard, Mathieu and Moga, Daniela C. and Quellec, Gwénolé and Niemeijer, Meindert},
    title = "{Automated Analysis of Retinal Images for Detection of Referable Diabetic Retinopathy}",
    journal = {JAMA Ophthalmology},
    volume = {131},
    number = {3},
    pages = {351-357},
    year = {2013},
    month = {03},
}

@article{mess2,
	author = {Etienne Decencière and Xiwei Zhang and Guy Cazuguel and Bruno Lay and Béatrice Cochener and Caroline Trone and Philippe Gain and Richard Ordonez and Pascale Massin and Ali Erginay and Béatrice Charton and Jean-Claude Klein},
	title = {FEEDBACK ON A PUBLICLY DISTRIBUTED IMAGE DATABASE: THE MESSIDOR DATABASE},
	journal = {Image Analysis \& Stereology},
	volume = {33},
	number = {3},
	year = {2014},
	issn = {1854-5165},	pages = {231--234}
}

@InProceedings{eccv20,
author="Liu, Jiazhen
and Li, Xirong
and Wei, Qijie
and Xu, Jie
and Ding, Dayong",
editor="Avidan, Shai
and Brostow, Gabriel
and Ciss{\'e}, Moustapha
and Farinella, Giovanni Maria
and Hassner, Tal",
title="Semi-supervised Keypoint Detector and Descriptor for Retinal Image Matching",
booktitle="Computer Vision -- ECCV 2022",
year="2022",
publisher="Springer",
pages="593--609",
isbn="978-3-031-19803-8"
}

@article{costeffec,
    author = {Ho, Ra AND Song, Lina D. AND Choi, Jin A. AND Jee, Donghyun},
    journal = {PLOS ONE},
    publisher = {Public Library of Science},
    title = {The cost-effectiveness of systematic screening for age-related macular degeneration in South Korea},
    year = {2018},
    month = {10},
    volume = {13},
    pages = {1-14},
    number = {10},

}

@article{rivas2,
	author = {David Rivas-Villar and Álvaro S. Hervella and José Rouco and Jorge Novo},
	title = {Joint keypoint detection and description network for color fundus image registration},
	journal = {Quantitative Imaging in Medicine and Surgery},
	volume = {},
	number = {},
	year = {},
	issn = {}
}

@InProceedings{fastap,
author = {Cakir, Fatih and He, Kun and Xia, Xide and Kulis, Brian and Sclaroff, Stan},
title = {Deep Metric Learning to Rank},
booktitle = {Proceedings of the IEEE/CVF Conference on Computer Vision and Pattern Recognition (CVPR)},
month = {June},
year = {2019}
}

@article{alvaro_asoc,
title = {Learning the retinal anatomy from scarce annotated data using self-supervised multimodal reconstruction},
journal = {Applied Soft Computing},
volume = {91},
pages = {106210},
year = {2020},
issn = {1568-4946},
author = {Álvaro S. Hervella and José Rouco and Jorge Novo and Marcos Ortega},
}

@article{rivas3,
  title={ConKeD: Multiview contrastive descriptor learning for keypoint-based retinal image registration},
  author={Rivas-Villar, David and Hervella, {\'A}lvaro S and Rouco, Jos{\'e} and Novo, Jorge},
  journal={arXiv:2401.05901},
  year={2024}
}

@InProceedings{kdsr,
    author    = {Nasser, Sahar Almahfouz and Gupte, Nihar and Sethi, Amit},
    title     = {Reverse Knowledge Distillation: Training a Large Model Using a Small One for Retinal Image Matching on Limited Data},
    booktitle = {Proceedings of the IEEE/CVF Winter Conference on Applications of Computer Vision (WACV)},
    month     = {January},
    year      = {2024},
    pages     = {7778-7787}
}

@InProceedings{geoformer,
    author    = {Liu, Jiazhen and Li, Xirong},
    title     = {Geometrized Transformer for Self-Supervised Homography Estimation},
    booktitle = {Proceedings of the IEEE/CVF International Conference on Computer Vision (ICCV)},
    month     = {October},
    year      = {2023},
    pages     = {9556-9565}
}

@inproceedings{loftr,
  title={LoFTR: Detector-free local feature matching with transformers},
  author={Sun, Jiaming and Shen, Zehong and Wang, Yuang and Bao, Hujun and Zhou, Xiaowei},
  booktitle={Proceedings of the IEEE/CVF conference on computer vision and pattern recognition},
  year={2021}
}

@article{retinaregnet,
title = {RetinaRegNet: A zero-shot approach for retinal image registration},
journal = {Computers in Biology and Medicine},
volume = {186},
pages = {109645},
year = {2025},
issn = {0010-4825},
author = {Vishal Balaji Sivaraman and Muhammad Imran and Qingyue Wei and Preethika Muralidharan and Michelle R. Tamplin and Isabella M. Grumbach and Randy H. Kardon and Jui-Kai Wang and Yuyin Zhou and Wei Shao}}

@inproceedings{gauss_primitive,
  title={Gaussian Primitive Optimized Deformable Retinal Image Registration},
  author={Tian, Xin and Wang, Jiazheng and Zhang, Yuxi and Chen, Xiang and Hu, Renjiu and Li, Gaolei and Liu, Min and Zhang, Hang},
  booktitle={International Conference on Medical Image Computing and Computer-Assisted Intervention},
  pages={218--228},
  year={2025},
  organization={Springer}
}

@INPROCEEDINGS{global_local,
  author={Liu, Yepeng and Yu, Baosheng and Chen, Tian and Gu, Yuliang and Du, Bo and Xu, Yongchao and Cheng, Jun},
  booktitle={2024 IEEE International Conference on Bioinformatics and Biomedicine (BIBM)}, 
  title={Progressive Retinal Image Registration via Global and Local Deformable Transformations}, 
  year={2024},
  volume={},
  number={},
  pages={2183-2190},
}

@InProceedings{censure,
author="Agrawal, Motilal
and Konolige, Kurt
and Blas, Morten Rufus",
editor="Forsyth, David
and Torr, Philip
and Zisserman, Andrew",
title="CenSurE: Center Surround Extremas for Realtime Feature Detection and Matching",
booktitle="Computer Vision -- ECCV 2008",
year="2008",
publisher="Springer Berlin Heidelberg",
address="Berlin, Heidelberg",
pages="102--115",
isbn="978-3-540-88693-8"
}

@InProceedings{fast,
author="Rosten, Edward
and Drummond, Tom",
editor="Leonardis, Ale{\v{s}}
and Bischof, Horst
and Pinz, Axel",
title="Machine Learning for High-Speed Corner Detection",
booktitle="ECCV 2006",
year="2006",
publisher="Springer Berlin Heidelberg",
address="Berlin, Heidelberg",
pages="430--443",
isbn="978-3-540-33833-8"
}

@inproceedings{harris,

   title = {A Combined Corner and Edge Detector},

   author = {Harris, C. and Stephens, M.},

   year = {1988},

   pages = {23.1-23.6},

   booktitle = {Proceedings of the Alvey Vision Conference},

   publisher = {Alvety Vision Club},

   editors = {Taylor, C. J.},

   isbn = {},

   note = {doi:10.5244/C.2.23}

}

@article{rivas4,
      title={ConKeD++ - Improving descriptor learning for retinal image registration: A comprehensive study of contrastive losses},
  author={Rivas-Villar, David and Hervella, {\'A}lvaro S and Rouco, Jos{\'e} and Novo, Jorge},
    journal={arXiv:2404.16773},
year={2024}
}

\end{document}